\def\year{2019}\relax
\documentclass[letterpaper]{article} 
\usepackage{aaai19}  
\usepackage{times}  
\usepackage{helvet}  
\usepackage{courier}  
\usepackage{url}  
\usepackage{graphicx}  
\usepackage{multirow}
\usepackage{bm}
\usepackage{amsfonts}
\usepackage{amsmath,amssymb}
\usepackage{algorithm}
\usepackage{algorithmic}
\usepackage{epstopdf}

\begin{document}
%
\title{Disentangled Variational Representation for Heterogeneous Face Recognition}

\author{
Xiang Wu$^{1, 2}$, Huaibo Huang$^{1, 2, 3}$, Vishal M. Patel$^{4}$, Ran He$^{1, 2}$\thanks{corresponding author}, Zhenan Sun$^{1, 2}$\\
$^{1}$ Center for Research on Intelligent Perception and Computing (CRIPAC), CASIA, Beijing, China\\
$^{2}$ National Laboratory of Pattern Recognition (NLPR), CASIA, Beijing, China\\
$^{3}$ School of Artificial Intelligence, University of Chinese Academy of Sciences, Beijing, China \\
$^{4}$Johns Hopkins University, 3400 N. Charles St, Baltimore, MD 21218, USA \\
{\tt\small alfredxiangwu@gmail.com, huaibo.huang@cripac.ia.ac.cn,} \\
{\tt\small vpatel36@jhu.edu, \{rhe, znsun\}@nlpr.ia.ac.cn}
}

\maketitle

\begin{abstract}
Visible (VIS) to near infrared (NIR) face matching is a challenging problem due to the significant domain discrepancy between the domains and a lack of sufficient data for training cross-modal matching algorithms. Existing approaches attempt to tackle this problem by either synthesizing visible faces from NIR faces, extracting domain-invariant features from these modalities, or projecting heterogeneous data onto a common latent space for cross-modal matching. In this paper, we take a different approach in which we make use of the Disentangled Variational Representation (DVR) for cross-modal matching. First, we model a face representation with an intrinsic identity information and its within-person variations. By exploring the disentangled latent variable space, a variational lower bound is employed to optimize the approximate posterior for NIR and VIS representations. Second, aiming at obtaining more compact and discriminative disentangled latent space, we impose a minimization of the identity information for the same subject and a relaxed correlation alignment constraint between the NIR and VIS modality variations. An alternative optimization scheme is proposed for the disentangled variational representation part and the heterogeneous face recognition network part. The mutual promotion between these two parts effectively reduces the NIR and VIS domain discrepancy and alleviates over-fitting. Extensive experiments on three challenging NIR-VIS heterogeneous face recognition databases demonstrate that the proposed method achieves significant improvements over the state-of-the-art methods.

\end{abstract}

\section{Introduction}
In recent years, methods based on deep convolution neural network (CNN) have shown impressive performance improvements for face detection and recognition problems \cite{Parkhi:2015,Wu2018ALC}. Despite the success of CNN-based methods in addressing various challenges in face recognition such as variations in pose, expression, aging, occlusion, disguise, and illumination, they are specifically designed to recognize face images that are collected at or near the visible (VIS) domain. However, in many real-world applications such as surveillance at night-time and in low-light conditions, one has to be able to recognize faces collected in thermal or near infrared (NIR) domains.  The performance of many CNN-based face recognition methods often degrades significantly when confronted by the NIR face images.  This is mainly due to the significant distributional change between the NIR and VIS domains.


\begin{figure}[t]
\centering

\includegraphics[width=0.46\textwidth]{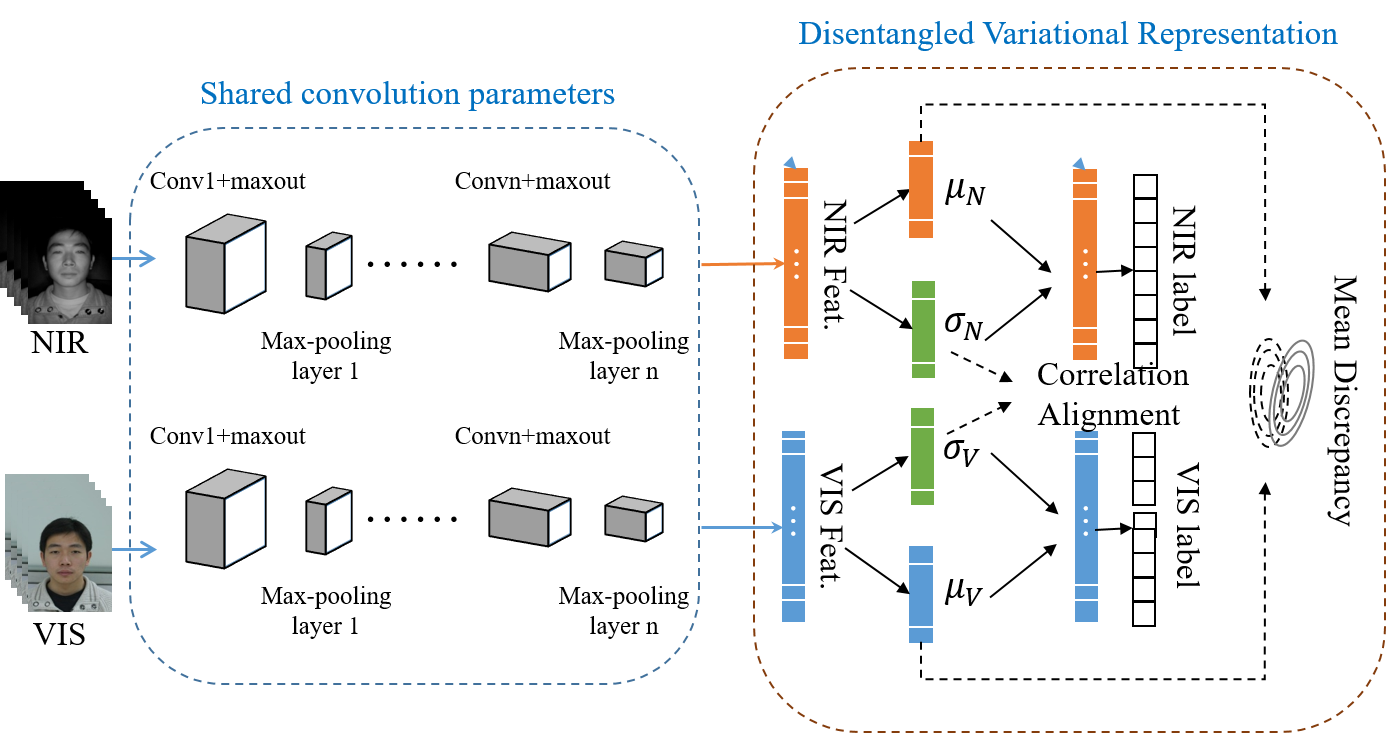}
\caption{An overview of the proposed Disentangled Variational Representation (DVR) approach for VIS-NIR matching. The NIR and VIS representations $x_N$ and $x_V$ are disentangled into ($\mu_N$, $\sigma_N$) and ($\mu_V$, $\sigma_V$), respectively. We assume that there is a linear relationship, $P$, between lighting variations, i.e., $\sigma_V=P\sigma_N$. The mean discrepancy is used to measure the difference between the NIR and VIS distributions in the latent  space. The reconstructions $\hat{x}_N$ and $\hat{x}_V$ are obtained from the likelihood $p(x_N|z_N)$ and $p(x_V|z_V)$, respectively and are constrained by the cross-entropy loss.}
\label{fig_framework}
\end{figure}

Another issue that one has to overcome when designing CNN-based models for heterogeneous face recognition (HFR) is over-fitting, which happens due to the lack of sufficient training samples.  One of the reasons why CNN-based face recognition methods provide impressive performance improvements on various face recognition benchmarks is that they are trained on thousands and millions of annotated face images often downloaded from the internet.  In contrast, there is no publicly available large-scale annotated NIR face dataset for training deep networks.  As a result, CNNs trained on small-scale NIR data often tend to overfit.  Hence, it is necessary to explore other methods that can deal with this issue in HFR.

Various methods have been developed in the literature for VIS to NIR cross-modal face recognition \cite{SLi:2013,Reale2016SeeingTF}.  In particular, methods such as  \cite{XXLiu:2016,RHe:2017,Patel_IJCB2017,XWu:2018,Song2018AdversarialDH,DBLP:journals/corr/abs-1708-02412,Patel_BTAS2018} attempt to reduce the domain gap between the NIR and VIS domains and learn domain invariant representations for HFR.

In contrast, we take a different approach in which we make use of the Disentangled Variational Representation (DVR) to deal with the two aforementioned challenges. First, inspired by the observation  that the facial appearance is composed of the identity information and the variation information, as shown in Fig.~\ref{fig_framework}, we assume that there exists an independent latent variable, which can be composed of an intrinsic variable for identity and an intra-personal variable for within-person variation. Second, benefiting from the variational lower bound~\cite{Kingma2013AutoEncodingVB} to tackle the marginal likelihood estimation, we model the approximate posterior and obtain disentangled latent variable. Next, when imposing the minimization of the identity information for the same subject and the assumption of correlation alignment \cite{BSun:2016} between different modality variations, we obtain more compact and discriminative disentangled latent space for DVR. Although there are large light spectrum variations, spectrum variations are often assumed to be on linear subspaces. Hence, we employ a relaxed correlation alignment item to constrain the variations of different modalities. Furthermore, generating samples from the approximate posterior significantly alleviates the need for having large number of samples during training the fully connected layers of deep HFR models. Since the effectiveness of the generated samples from the likelihood depends on the estimated approximate posterior, we propose  an alternative optimization approach for the DVR framework during training in which HFR network can contribute to the disentangled representation training and vice versa.

To summarize, the following are our main contributions:
\begin{itemize}
  \item An end-to-end DVR framework is developed for cross-modal NIR-VIS face matching. We introduce a variational lower bound to estimate the posterior and optimize the latent variable space, aiming at disentangling the NIR and VIS face representations.
  \item We propose to minimize the identity information for the same subject and the relaxed correlation alignment constraint on modality variations that facilitate modeling the compact and discriminative disentangled latent variable spaces for heterogeneous modalities.
  \item An alternative optimization is proposed to provide mutual promotion between HFR network and disentangled variational representation part. Thus, DVR can both reduce the domain discrepancy and alleviate over-fitting.
  \item Extensive experimental results are conducted
  on three HFR databases, including the CASIA NIR-VIS 2.0 database \cite{SLi:2013}, the Oulu-CASIA NIR-VIS database \cite{JChen:2009} and the BUAA-VisNir database \cite{DHuang:2012}, and comparisons are performed against several recent state-of-the-art approaches. Furthermore, an ablation study is conducted to demonstrate the improvements obtained by various components of the proposed method.

\end{itemize}

\section{Related Work}
We follow the notations in \cite{Zhu2014MatchingNF,RHe:2017,DBLP:journals/corr/abs-1708-02412,Song2018AdversarialDH} while providing a brief survey of HFR and disentangled representation learning.
\subsection{Heterogeneous Face Recognition (HFR)}
The problem of HFR has gained a lot of traction in recent years  \cite{Xiao2013CoupledFS,Ouyang2016ASO}.
 According to~\cite{Zhu2014MatchingNF}, the existing methods are divided into the following three main categories:

\textbf{Latent subspace learning} aims to project the heterogenous data onto a common latent space in which the relevance of heterogeneous data can be measured.  Lin~\cite{Lin2006IntermodalityFR} proposed a Common Discriminant Feature Extraction (CDFE) method to incorporate both discriminative and locality information. By introducing feature selection via nuclear norm, a common subspace learning was employed in~\cite{Wang2016JointFS}. Shao \emph{et al}~\cite{Shao2014GeneralizedTS} project NIR and VIS data into a generalized subspace where each NIR sample can be represented by a combination of VIS samples. Restricted Boltzmann Machines (RBMs) are employed in ~\cite{Yi2014SharedRL} to learn a shared representation between different domains and then Principal Component Analysis (PCA) is applied to remove the redundancy and heterogeneity. Wang \emph{et al}~\cite{Wang2015OnDM} propose several deep neural network-based methods with Canonical Correlation Analysis (CCA) in unsupervised subspace feature learning for HFR. He \emph{et al}~\cite{RHe:2017,DBLP:journals/corr/abs-1708-02412} divide the high-level representation into two orthogonal subspaces to obtain domain-invariant identity information and domain-related spectrum information.

\textbf{Modality-invariant feature learning} explores domain-invariant features that are only related to the face identity. Traditional methods are based on the handcrafted local features~\cite{Liao2009HeterogeneousFR,Klare2011MatchingFS,Goswami2011EvaluationOF}, including Local Binary Patterns (LBP), Gabor features~\cite{ZLei:2007}, Histograms of Oriented Gradients (HOG) and Difference of Gaussian (DoG). Liao \emph{et al}~\cite{Liao2009HeterogeneousFR} combine DoG filtering and multi-block LBP to encode NIR and VIS images. Klare \emph{et al}~\cite{Klare2011MatchingFS} utilize HOG features with sparse representation to improve the performance of HFR. Goswami \emph{et al}~\cite{Goswami2011EvaluationOF} combine the LBP histogram representation with Linear Discriminant Analysis (LDA) to extract domain invariant features. As for deep learning, Kan \emph{et al}~\cite{Kan2016MultiviewDN} address the discriminant domain invariant feature learning by analyzing the within-class and between-class scatter. Coupled Deep Learning (CDL)~\cite{XWu:2018} utilizes nuclear norm constraint on fully connected layer to alleviate overfitting, and proposes a cross-modal ranking to reduce domain discrepancy. He \emph{et al}~\cite{DBLP:journals/corr/abs-1708-02412} decrease the domain gap by Wasserstein distance to obtain domain invariant features for HFR.

\textbf{Data synthesis} attempts to address the domain discrepancy at image level by  transforming face images from one modality into another via image synthesis. Data synthesis is first proposed to synthesize and recognize a sketch image from a face photo in~\cite{Tang2003FaceSS}. Wang~\cite{Wang2009FacePS} applies Markov Random Field (MRF) to transform pseudo-sketch to face photo in a multi-scale way. In~\cite{JuefeiXu2015NIRVISHF}, joint dictionary learning is used to reconstruct face images and then perform face matching. Lezama \emph{et al}~\cite{Lezama2017NotAO} propose a cross-spectral hallucination and low-rank embedding to synthesize a heterogeneous image in a patch way. With developments of a photo-realistic synthesis image by Generative Adversarial Network (GAN)~\cite{Goodfellow2014GenerativeAN}, ``recognition via generation''~\cite{Zhao2017DualAgentGF,Huang2017BeyondFR,hu2018pose} is drawn attention by lots of researchers. Song \emph{et al}~\cite{Song2018AdversarialDH} utilize a Cycle-GAN~\cite{Zhu2017UnpairedIT} to realize a cross-spectral face hallucination, facilitating heterogeneous face recognition via generation. However, due to the small number of images in the training set, there are still challenges to synthesize photo-realistic VIS face images from NIR images.

\subsection{Learning to Disentangled Representations}
Early work~\cite{Schmidhuber1992LearningFC} attempts to disentangle representations in an autoencoder via penalizing predictability of latent variables. A variant of Boltzmann Machine~\cite{Desjardins2012DisentanglingFO} is used to disentangle factors of variations in the training data. Kingma~\cite{Kingma2013AutoEncodingVB} propose the Variational Auto-Encoder (VAE) framework to achieve limited disentangling performance on simple datasets. Matthey \emph{et al}~\cite{Matthey2016vaeLB} augment the original VAE framework with a single hyper-parameter $\beta$, called $\beta$-VAE, that controls the degree of disentanglement in the latent representations. Besides, FactorVAE~\cite{Kim2018DisentanglingBF} is proposed to disentangle by encouraging the distribution of representation to be factorial and independent across the dimensions.  Chen~\cite{Chen2012BayesianFR} propose joint Bayesian formulation to decompose a face representation into three parts, including intrinsic difference, transformation difference and noise. An expectation maximization-like learning procedure is employed to optimize the joint formulation and they achieve promising performance on the face recognition tasks. Shi \emph{et al}~\cite{Shi2017CrossModalityFR} extend the original joint Bayesian approach by modeling the gallery and probe images using two different Gaussian distributions to propose a heterogeneous joint Bayesian approach for HFR.

\section{Proposed Method}
We begin this section by reviewing the Wasserstein CNN method~\cite{DBLP:journals/corr/abs-1708-02412} that introduces a probabilistic framework for HFR and shows promising results. Based on the Wasserstein CNN, we give the details of our disentangled variational representation method and the corresponding optimization scheme.
\subsection{Revisiting Wasserstein CNN}
Let $x_N \in \mathbb{R}^d$ and $x_V \in \mathbb{R}^d$ denote the NIR and VIS domain data representations, respectively. In Wasserstein CNN~\cite{DBLP:journals/corr/abs-1708-02412}, it is assumed that the data distributions of the representations for the same identity follow a Gaussian distribution. Hence,  $x_N\sim \mathcal{N}(m_N, C_N)$ and $x_V\sim \mathcal{N}(m_V, C_V)$, where $m_N, m_V$ are the mean vectors and $C_N, C_V$ are the covariance matrices. The 2-Wasserstein distance between $x_N$ and $x_V$ corresponding to the same identity is defined as
\begin{align}
   \nonumber W(x_N, x_V)&=\|m_N-m_V\|^2_2 \\&+ \text{trace}(C_N+C_V-2(C_V^{\frac{1}{2}}C_NC_V^{\frac{1}{2}})).
\label{eq_w_distance}
\end{align}
Due to the ability of measuring the consistency between two distributions, Eq. (\ref{eq_w_distance}) is used to reduce the domain gap between the NIR and VIS images. However, the Wasserstein distance is directly imposed on the NIR and VIS representations, which are obtained from a CNN. It is well-known that CNN-based NIR and VIS representations contain various  high-level information including identity, spectrum, pose, noise, etc., which are not disentangled. Therefore, directly matching representation distributions may not lead to better performance especially when the training set is not large enough for HFR.

\subsection{Disentangled Variational Representation}\label{sec_dvr}
Let $\{x^{(i)}\in\mathbb{R}^d \}_{i=1}^{N}$ and $\{z^{(i)}\in\mathbb{R}^h \}_{i=1}^{N}$ denote $N$ observations and the independent latent variables corresponding to one identity, respectively.  For each sample $x^{(i)}$, we can obtain
\begin{equation}
    z^{(i)}=\mu^{(i)} + \epsilon\odot\sigma^{(i)},
    \label{eq_sample}
\end{equation}
where $\mu^{(i)}$ represents the identity information,  $\sigma^{(i)}$ contains variations, $\epsilon\sim \mathcal{N}(\bm{0},{\text{I}})$, $\mu^{(i)}, \sigma^{(i)}, \epsilon \in \mathbb{R}^d$, and $\odot$ denotes the Hadamard product.  Note that the marginal likelihood $p(x)=\int p(x|z)p(z)$ is intractable. Hence, different from the common simplifying assumptions about the marginal or posterior probabilities, we introduce the variational lower bound (or evidence lower bound, ELBO)
\begin{align}
   \nonumber  \log p(x^{(i)}) &\geq -\text{KL}(q(z|x^{(i)})||p(z)) \\&+ \mathbb{E}_{q(z|x^{(i)})}\left[\log p(x^{(i)}|z)\right],
    \label{eq_vae}
\end{align}
where $q_{\phi}(z|x^{(i)})$ can be implemented by a probabilistic encoder, $q_{\phi}(z|x^{(i)})\sim \mathcal{N}(z; {\mu}^{(i)}, {\sigma}^{2(i)}{\text{I}})$, and $\phi$ denotes the parameters.  Note that the posterior $p(x^{(i)}|z)$ can be treated as the reconstruction part. Let the prior over the latent variables $z$ be a centered isotropic multivariate Gaussian $p(z)\sim \mathcal{N}(\bm{0},{\text{I}})$.  As a result, the disentangled formulation in Eq. (\ref{eq_vae}) can be treated as a variational autoencoder~\cite{Kingma2013AutoEncodingVB}.

Let $z_N \in \mathbb{R}^h, z_V \in \mathbb{R}^h$ represent the latent variables corresponding to the NIR and VIS representations $x_N \in \mathbb{R}^d, x_V \in \mathbb{R}^d$, respectively.  Then, one can approximate the posterior as follows:
\begin{equation}
\begin{array}{l}
    q_N(z_N|x_N^{(i)})\sim \mathcal{N}(z_N; {\mu}_N^{(i)}, {\sigma}_N^{2(i)}{\text{I}})\\
    q_V(z_V|x_V^{(i)})\sim \mathcal{N}(z_V; {\mu}_V^{(i)}, {\sigma}_V^{2(i)}{\text{I}}),
\end{array}
\label{eq_prob_ae}
\end{equation}
where $z_N=\mu_N+\epsilon\odot\sigma_N$, $z_V=\mu_V+\epsilon\odot\sigma_V$ and $\epsilon\sim \mathcal{N}(\bm{0},{\text{I}})$. Here, $\phi_N$ and $\phi_V$ denote the parameters of the NIR and VIS approximate posterior estimator, respectively. In NIR-to-VIS face recognition, the main discrepancy comes from the variation in the light spectrum of NIR and VIS domains.  We assume that the light spectrum variations are related as follows
\begin{equation}
    \sigma_V = P\sigma_N,
    \label{eq_linear_projection}
\end{equation}
where $P\in \mathbb{R}^{h\times h}$ is a correlation alignment matrix. Different from \cite{BSun:2016}, we assume that there is a linear relationship between covariance matrices rather than requiring them to be similar. Since the latent variables $z_N$ and $z_V$ are independent, we impose an orthogonality constraint on $P$. Therefore, Eq. (\ref{eq_prob_ae}) can be reformulated as
\begin{equation}
\begin{array}{c}
    q_N(z_N|x_N^{(i)})\sim \mathcal{N}(z_N; {\mu}_N^{(i)}, {\sigma}_N^{2(i)}{\text{I}})\\
    q_V(z_V|x_V^{(i)})\sim \mathcal{N}(z_V; {\mu}_V^{(i)}, {\sigma}_V^{2(i)}{\text{I}}) \\
    \text{s.t. } \sigma_V = P\sigma_N, \quad P^\top P=I.
\end{array}
\end{equation}
The correlation alignment matrix $P$ plays the role of constraining the variations of $\sigma_N$ and $\sigma_V$. It makes the representations of NIR and VIS images vary in a subspace. Experimental results also verify the effectiveness of this correlation alignment constraint. Furthermore, since $\mu_N$ and $\mu_V$ represent the identity information, benefiting from the Wasserstein CNN, we minimize $\|\mu_N-\mu_V\|^2_2$ for the same identities to reduce the domain discrepancy.

With the above definitions, the proposed DVR formulation is as follows
\begin{equation}
\resizebox{0.49\textwidth}{!}{$
\begin{array}{cl}
\mathcal{J}_{\text{DVR}} = & -\underbrace{\left[\text{KL}(q(z_N|x_N^{(i)})||p(z_N)) +\text{KL}(q(z_V|x_V^{(i)})||p(z_V)) \right]}_{\text{approximate posterior estimator parts}}\\
     & + \underbrace{\mathbb{E}\left[\log p(x_N^{(i)}|z_N)\right] + \mathbb{E}\left[\log p(x_V^{(i)}|z_V)\right]}_{\text{reconstruction parts}}\\
     & +\underbrace{\lambda_1\|\mu^{(i)}_N-\mu^{(i)}_V\|^2_2}_{\text{mean discrepancy part}} \\
     & \qquad\qquad \text{s.t. } \sigma_V = P\sigma_N, \quad P^\top P=I.
\end{array} $}
\label{eq_loss_dvr}
\end{equation}
Using the Lagrange multipliers and the reparameterization trick~\cite{Kingma2013AutoEncodingVB}, Eq. (\ref{eq_loss_dvr}) can be reformulated as
\begin{equation}
    \begin{array}{cl}
    \mathcal{J}_{\text{DVR}} = & -\frac{1}{2}\underbrace{\sum_{j}\left(1+\log\sigma_{Nj}^{2(i)}-\mu_{Nj}^{2(i)}-\sigma_{Nj}^{2(i)}\right)}_{\text{NIR approximate posterior estimator}} \\
         & - \frac{1}{2}\underbrace{\sum_{j}\left(1+\log\sigma_{Vj}^{2(i)}-\mu_{Vj}^{2(i)}-\sigma_{Vj}^{2(i)}\right)}_{\text{VIS approximate posterior estimator}} \\
         &  + \underbrace{\mathbb{E}\left[\log p(x_N^{(i)}|z_N)\right] + \mathbb{E}\left[\log p(x_V^{(i)}|z_V)\right]}_{\text{reconstruction parts}}\\
         & +\underbrace{\lambda_1\|\mu^{(i)}_N-\mu^{(i)}_V\|^2_2}_{\text{mean discrepancy part}} \\
         & + \underbrace{\lambda_2\|\sigma_V - P\sigma_N\|^2_2 + \lambda_3\|P^\top P -I\|^2_F}_{\text{correlation alignment constraint}},
    \end{array}
    \label{eq_reformuate_loss_dvr}
\end{equation}
where $j$ denotes the $j$-th element of vectors $\mu^{(i)}_N, \mu^{(i)}_V, \sigma^{(i)}_N$ and $\sigma^{(i)}_V$, $\|\cdot\|^2_F$ denotes the Frobenius norm, and $\lambda_1, \lambda_2$ are the trade-off parameters.

As for the reconstruction parts, we let $p(x|z)$ be a multivariate Gaussian which is computed from $z$ with a multilayer perceptron (MLP). Therefore, given $x^{(i)}_N$ and $x^{(i)}_V$ with the label $y$, we can generate $\hat{x}^{(i)}_N$ and $\hat{x}^{(i)}_V$ from the likelihood $p(x_N^{(i)}|z_N)$ and $p(x_V^{(i)}|z_V)$, respectively.
In ~\cite{Kingma2013AutoEncodingVB}, the L2 losses $\|x^{(i)}_N-\hat{x}^{(i)}_N\|^2_2$ and  $\|x^{(i)}_V-\hat{x}^{(i)}_V\|^2_2$ are used for the reconstruction parts. In DVR, except for the L2 reconstruction loss, we further impose the cross-entropy loss between the reconstructions $\hat{x}^{(i)}_N$ and $\hat{x}^{(i)}_V$ sampled from the posteriors and the identity label $y$.

\subsection{Heterogeneous Recognition Network}
Given NIR and VIS face images, $I_N$ and $I_V$ respectively, we  denote the CNN features as $x_i=f(I_i; \Theta), i\in\{N, V\}$. The output of a CNN feature is normally fed into a softmax layer for supervised training,
\begin{equation}
    \mathcal{J}_{\text{cls}}=\text{softmax}(x_i;W, \Theta), i\in\{N, V\}.
\end{equation}
Given a training sample $(x_i, y), i\in\{N,V\}$,  we can generate $(\hat{x}_i, y), i\in\{N,V\}$ using the DVR framework, which can also be fed into a softmax layer as follows
\begin{equation}
\resizebox{0.45\textwidth}{!}{
    $\mathcal{J}_{\text{cls}}=\text{softmax}(x_i, y;W, \Theta) + \text{softmax}(\hat{x}_i, y;W, \Theta), i\in\{N, V\}$ }.
    \label{eq_softmax}
\end{equation}
On the one hand, benefiting from the generated samples $\hat{x}_N$ and $\hat{x}_V$, the CNN feature extraction part $f(\cdot;\Theta)$ can be better optimized and more robust, especially when the training sets for HFR are not large enough. On the other hand, the more robust the CNN feature extraction $f(\cdot;\Theta)$ is, the more precisely the approximate posteriors $q(z_i|x_i), i\in\{N, V\}$ in DVR can be estimated. Inspired by this assumption, we propose an alternative optimization method to obtain domain-invariant representations for HFR.

\begin{algorithm}[t]
\caption{Disentangled Variational Representation (DVR) Training.}
\begin{algorithmic}[1]
\REQUIRE
Training set: NIR images $I_N$, VIS images $I_V$, the learning rate $\alpha$ and the trade-off parameters $\lambda_1, \lambda_2, \lambda_3$.
\ENSURE
The CNN parameters $\Theta, W$, the approximate posterior estimators $\phi_N, \phi_V$ and correlation alignment matrix $P$.
\STATE Initialize $\Theta, W$ by pre-trained model;
\STATE Obtain $x_N=f(I_N;\Theta), x_V=f(I_V;\Theta)$;
\STATE Initialize $\phi_N, \phi_V, P$ randomly;
\FOR {$t=1,\dots, T$}
\STATE Optimize $\phi_N, \phi_V$ without mean discrepancy and correlation alignment parts;
\ENDFOR;
\FOR {$t=1,\dots, T$}
\STATE Given $\epsilon\sim \mathcal{N}(\bm{0},{\text{I}})$, generate $\hat{x}_N$ and $\hat{x}_V$ via Eq. (\ref{eq_sample}) and Eq. (\ref{eq_prob_ae});
\STATE Compute loss $\mathcal{J}_{\text{cls}}$ via Eq. (\ref{eq_softmax})
\STATE Fix $\phi_N, \phi_V, P$;
\STATE \quad Update $\Theta, W$ via back-propagation;
\STATE Obtain $x_N=f(I_N;\Theta), x_V=f(I_V;\Theta)$;
\STATE Fix $\Theta, W, P$
\STATE \quad Update $\phi_N, \phi_V$ by Eq. (\ref{eq_reformuate_loss_dvr});
\STATE Fix $\Theta, W, \phi_N, \phi_V$
\STATE \quad Update $P$ by gradient descent;
\ENDFOR;
\STATE \textbf{Return} $\Theta, W, \phi_N, \phi_V, P$;
\end{algorithmic}
\label{alg1}
\end{algorithm}

\subsection{Optimization}

In this section, we present an alternative optimization method for the DVR framework. The CNN feature extraction part $f(\cdot;\Theta)$ is initialized by a pre-trained model. First, we directly optimize the approximate posteriors $q(z_i|x_i), i\in\{N, V\}$ until convergence by Eq. (\ref{eq_reformuate_loss_dvr}), but without mean discrepancy and correlation alignment parts, from the random initialization. Second, we generate $\hat{x}_N$ and $\hat{x}_V$ according to Eq. (\ref{eq_sample}) and Eq. (\ref{eq_prob_ae}). We then fix the parameters $\phi_N$ and $\phi_V$ in approximate posterior estimator parts and compute Eq. (\ref{eq_softmax}) as the loss function to optimize the parameters of the recognition network $\Theta$ and $W$. Finally, the parameters $\Theta$ and $W$ in the recognition network are fixed. We utilize the output $x_i=f(I_i;\Theta), i\in\{N, V\}$ as the input, which contributes to the optimization of the approximate posterior estimator parts. The optimization details are summarized in Algorithm \ref{alg1}.

Regarding testing for HFR, we directly employ the outputs of heterogeneous recognition network $f(\cdot;\Theta)$ to obtain $x_i (i\in\{N, V\})$ as feature representations. The cosine distance is used to compute the similarity score between different heterogenous representations for evaluations. Note that the parameters $\phi_N, \phi_V, P$ of disentangled variational part and correlation alignment part are not utilized for testing. These two parts aim to disentangle representations and play a role of regularization; therefore, they are only utilized for training to reduce the domain discrepancy and alleviate overfitting. Experimental results demonstrate that these two parts can facilitate convolutional layers to learn a better feature representation.


\section{Experimental Results}
In this section, the proposed variational representation learning framework is systemically evaluated against several state-of-the-art HFR methods. We follow the experimental settings proposed in \cite{RHe:2017}\cite{XWu:2018}\cite{Song2018AdversarialDH} and mainly employ NIR and VIS images to perform experiments. Both quantitative results and qualitative results are reported.
\subsection{Datasets and Protocols}
Three publicly available VIS-to-NIR face recognition datasets are used to evaluate the performance of different HFR methods.


\textbf{The CASIA NIR-VIS 2.0 Face Database}~\cite{SLi:2013} is the largest and most challenging NIR-VIS heterogeneous face recognition database due to the large variations in lighting, expression and pose. It consists of 725 identities, each with 1 to 22 VIS and 5 to 50 NIR images. It consists of 10-fold experiments. For training, there are about 2,500 VIS and 6,100 NIR images from 360 identities. For testing, the gallery set in each fold is constructed from 358 identities and each identity only has one VIS image. The probe set contains over 6,000 NIR images from the same 358 identities. All the NIR images in the probe set are to be matched against the VIS images in the gallery set, resulting in a $6000\times358$ similarity matrix. The Rank-1 accuracy and verification rate (VR)@ false accept rate (FAR)=0.1\% are reported.

\textbf{The Oulu-CASIA NIR-VIS Database}~\cite{JChen:2009} contains 80 identities with 6 expression variations. Following the protocols in~\cite{DBLP:journals/corr/abs-1708-02412}, we select 20 identities as the training set and 20 identities as the testing set. Eight face images from each expression are randomly selected from both NIR and VIS sets. Hence, there are totally 96 images per each subject. All the VIS images of the 20 subjects are used as the gallery set and all the NIR images are treated as the probe set. The similarity matrix between the probe set and the gallery set is of size $960\times960$. The rank-1 accuracy, VR@FAR=1\% and VR@FAR=0.1\% are reported for comparisons.

\textbf{The BUAA-VisNir Face Database}~\cite{DHuang:2012} consists of data from 150 subjects with 9 VIS and 9 NIR face images per subject. The training set and testing set are composed of 900 images from 50 identities and 1800 images from the remaining 100 identities, respectively. Only one VIS image is selected in the gallery set and the probe set contains 900 NIR images during testing. The similarity matrix between the probe set and the gallery set is of size $900\times100$.  The rank-1 accuracy, VR@FAR=1\% and VR@FAR=0.1\% are reported for comparisons.

\begin{table*}[t]
\centering
\caption{The ablation study for DVR. Both LightCNN-9 and LightCNN-29 are used as the backbones. }
\resizebox{0.9\textwidth}{!} {
\begin{tabular}{|c|ccc|cc|ccc|ccc|}
\hline
\multirow{2}{*}{Backbone} & Disentangled & Mean & Correlation  &  \multicolumn{2}{c|}{CASIA NIR-VIS 2.0} & \multicolumn{3}{c|}{Oulu-CASIA NIR-VIS}  & \multicolumn{3}{c|}{BUAA-VisNir} \\ \cline{5-12}
& Variational Part & Discrepancy & Alignment &Rank-1 & FAR=0.1\% & Rank-1 & FAR=1\% & FAR=0.1\% & Rank-1 & FAR=1\%&  FAR=0.1\% \\
\hline
\multirow{4}{*}{LightCNN-9} & -&- &- & $97.1$ & $93.7$ & $93.8$ & $80.4$ & $43.8$ & $94.8$ & $94.3$ & $83.5$\\
& $\surd$ & - & - & $98.0$ & $97.3$ & $96.3$ & $85.9$ & $50.7$ & $96.5$ & $95.8$ & $88.3$\\
& $\surd$ &  $\surd$ & - & $98.2$ & $98.1$ & $98.0$ & $88.6$ & $61.3$ & $97.3$ & $96.6$ & $91.0$\\
& $\surd$ &  $\surd$ & $\surd$ & \bm{$99.1$} & \bm{$98.6$} & \bm{$99.3$} & \bm{$89.7$} & \bm{$65.8$} & \bm{$97.9$} & \bm{$97.0$} & \bm{$92.8$}\\
\hline
\multirow{4}{*}{LightCNN-29} &- & - & - & $98.1$ & $97.4$ & $99.0$ & $93.1$ & $68.3$ & $96.8$ & $97.0$ & $89.4$ \\
& $\surd$ &  -& - & $99.0$ & $99.1$ & \bm{$100.0$} & $95.2$ & $79.8$ & $98.0$ & $97.9$ & $93.0$\\
& $\surd$ &  $\surd$ & - & $99.5$ & $99.3$ &  \bm{$100.0$} & $96.5$ & $83.0$ & $98.9$ & $98.4$ & $95.6$\\
& $\surd$ &  $\surd$ & $\surd$ & \bm{$99.7$} & \bm{$99.6$} & \bm{$100.0$} & \bm{$97.2$} & \bm{$84.9$} & \bm{$99.2$} & \bm{$98.5$}& \bm{$96.9$} \\

\hline
\end{tabular}
}
\label{tab:ablation_study}
\end{table*}

\subsection{Implementation Details}

We employ the Light CNN~\cite{Wu2018ALC} as a basic network architecture for HFR. Both LightCNN-9 and LightCNN-29 models\footnote{https://github.com/AlfredXiangWu/LightCNN} are used as the backbone networks, which are pre-trained on the MS-Celeb-1M dataset~\cite{DBLP:journals/corr/GuoZHHG16}. All the images in the  training set are aligned to $144\times144$ and randomly cropped to $128\times128$ as the input. Stochastic gradient descent (SGD) is used, where the momentum is set to 0.9 and weight decay is set to $5e$-$4$. The learning rate is set to $1e$-$4$ initially and reduced to $5e$-$5$ gradually. The batch size is set to 128 and the dropout ratio is 0.5.

A multilayer perceptron (MLP) is used to model the DVR parts. It contains four hidden layers with $h$ dimensions to represent $\mu_N$, $\mu_V$, $\sigma_N$ and $\sigma_V$. Moreover, the correlation alignment matrix $P$ is an $h\times h$ matrix. Specifically, in the experiments, the dimension $h$ is set equal to $64$. The input and the output layers are both $256$-d, which are similar to the dimensions of features from the face recognition network. During training, the parameters of MLP are initialized by a Gaussian, while $P$ is initialized by an identity matrix $I$. Adam~\cite{Kingma2014AdamAM} is used for back-propagation and the initial learning rate is set $1e$-$3$ and gradually reduced to $1e$-$5$. The batch size is set to 128. The trade-off parameters $\lambda_1, \lambda_2$ and $\lambda_3$ are set equal to $1.0$, $0.1$ and $0.001$, respectively.

\begin{table*}[ht]
\centering
\caption{Comparisons with other state-of-the-art HFR methods on the CASIA NIR-VIS 2.0 database, the Oulu-CASIA NIR-VIS database and the BUAA-VisNir database.}
\resizebox{0.9\textwidth}{!} {
\begin{tabular}{|c|cc|ccc|ccc|}
\hline
\multirow{2}{*}{Method} &  \multicolumn{2}{c|}{CASIA NIR-VIS 2.0} & \multicolumn{3}{c|}{Oulu-CASIA NIR-VIS}  & \multicolumn{3}{c|}{BUAA-VisNir} \\ \cline{2-9}
& Rank-1 & FAR=0.1\% & Rank-1 & FAR=1\% & FAR=0.1\% & Rank-1 & FAR=1\%&  FAR=0.1\% \\
\hline
KDSR~\cite{Huang2013RegularizedDS} & $37.5$ & $9.3$ & $66.9$ & $56.1$ & $31.9$ & $83.0$ & $86.8$ & $69.5$  \\
H2(LBP3)~\cite{Shao2017CrossModalityFL} & $43.8$ & $10.1$ & $70.8$ & $62.0$ & $33.6$ & $88.8$ & $88.8$ & $73.4$ \\
Gabor+RBM~\cite{Yi2014SharedRL} & $86.2\pm1.0$ & $81.3\pm1.8$  & - & - & - & - & - & -\\
Recon.+UDP~\cite{JuefeiXu2015NIRVISHF} & $78.5\pm1.7$ & $85.8$ & - & - & - & - & - & -\\
Gabor+JB~\cite{Chen2012BayesianFR} & $89.5\pm0.8$ & $83.2\pm1.0$ & - & - & - & - & - & -\\
Gabor+HJB~\cite{Shi2017CrossModalityFR} & $91.6\pm0.8$ & $89.9\pm0.9$ & - & - & - & - & - & -\\
\hline
IDNet~\cite{Reale2016SeeingTF} & $87.1\pm0.9$ & $74.5$ & - & - & - & - & - & -\\
HFR-CNN~\cite{Saxena2016HeterogeneousFR} & $85.9\pm0.9$ & $78.0$ & - & - & - & - & - & -\\
Hallucination~\cite{Lezama2017NotAO} & $89.6\pm0.9$ & - & - & - & - & - & - & -  \\
TRIVET~\cite{XXLiu:2016} & $95.7\pm0.5$ & $91.0\pm1.3$ & $92.2$ & $67.9$ & $33.6$ & $93.9$ & $93.0$ & $80.9$\\
IDR~\cite{RHe:2017} & $97.3\pm0.4$ & $95.7\pm0.7$ & $94.3$ & $73.4$ & $46.2$ & $94.3$ & $93.4$ & $84.7$\\
ADFL~\cite{Song2018AdversarialDH} & $98.2\pm0.3$ & $97.2\pm0.3$ & $95.5$ & $83.0$ & $60.7$ & $95.2$ & $95.3$ & $88.0$ \\
CDL~\cite{XWu:2018} & $98.6\pm0.2$ & $98.3\pm0.1$ &  $94.3$ & $81.6$ & $53.9$ & $96.9$ & $95.9$ & $90.1$\\
W-CNN~\cite{DBLP:journals/corr/abs-1708-02412} & $98.7\pm0.3$ & $98.4\pm0.4$ &  $98.0$ & $81.5$ & $54.6$ & $97.4$ & $96.0$ & $91.9$\\
\hline
DVR (LightCNN-9) & $99.1\pm0.2$ & $98.6\pm0.2$ & $99.3$ & $89.7$ & $65.8$ & $97.9$ & $97.0$ & $92.8$ \\
DVR (LightCNN-29) & \bm{$99.7\pm0.1$} & \bm{$99.6\pm0.3$} & \bm{$100.0$} & \bm{$97.2$} & \bm{$84.9$} & \bm{$99.2$} & \bm{$98.5$}& \bm{$96.9$} \\
\hline
\end{tabular}
}
\label{tab:HFR}
\end{table*}

\begin{figure*}[ht]
\centering
\includegraphics[width=0.85\textwidth]{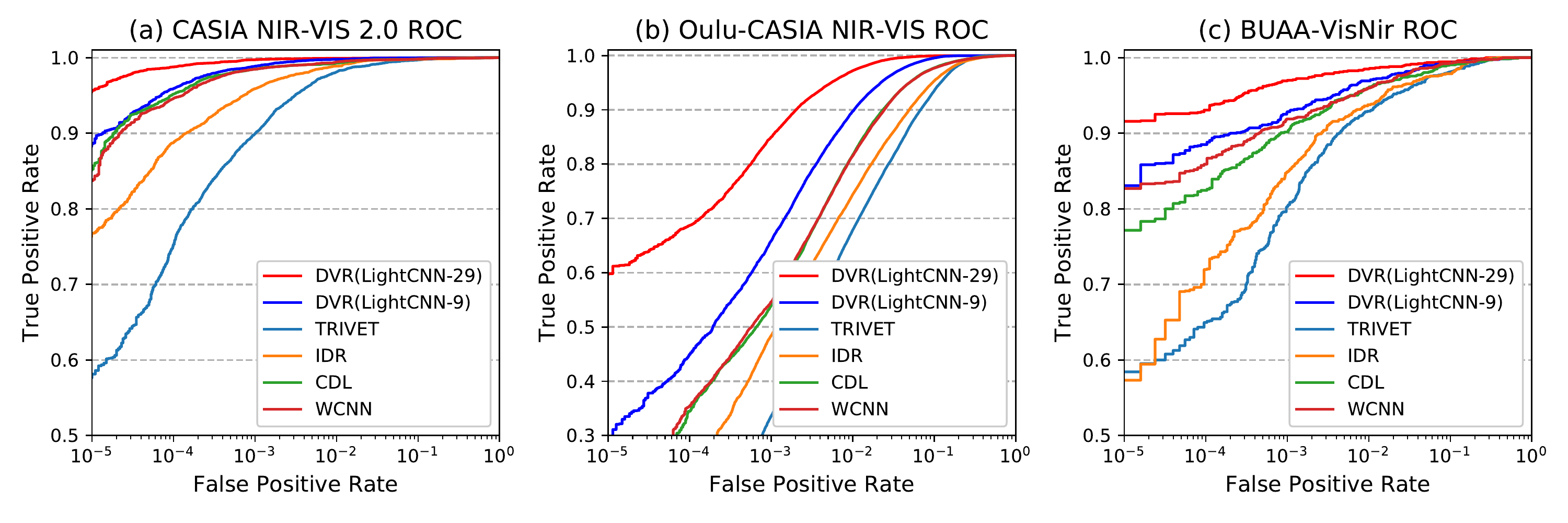}
\caption{The ROC curves on the CASIA NIR-VIS 2.0, the Oulu-CASIA NIR-VIS and the BUAA-VisNir databases, respectively}
\label{roc}
\end{figure*}

\subsection{Analysis of the Proposed Method\label{sec:ana}}
We first compare the performance between LightCNN-9 and LightCNN-29 models since they are used as the backbones for HFR. As shown in Table \ref{tab:ablation_study}, the LightCNN-29 achieves better performance than LightCNN-9 on all the three HFR databases.  This clearly shows that LightCNN-29 is more suitable and robust as a backbone network for HFR.

The aim of the proposed DVR is to model the disentangled latent variables $z_N$ and $z_V$ via $q(z_N|x_N)$ and $q(z_V|x_V)$ for NIR and VIS representations $x_N$ and $x_V$, respectively. And then, we can easily sample $\hat{x}_N$ and $\hat{x}_V$ according to the likelihood $p(x_N|z_N)$ and $p(x_V|z_V)$. Table \ref{tab:ablation_study} presents that on the Oulu-CAISA NIR-VIS database, with the disentangled variational part, the performance on VR@FAR=0.1\% is improved from 43.8\% to 50.7\% for LightCNN-9 and from 68.3\% to 79.8\% for LigthCNN-29, respectively. The results indicate that DVR can alleviate the lack of training data for HFR.

Similar with the Wasserstein CNN, minimizing mean discrepancy on $\mu_N$ and $\mu_V$ can significantly reduce the domain gap, which achieves \bm{$0.8\%$}, \bm{$10.6\%$} and \bm{$2.7\%$} improvements on VR@FAR=0.1\% with LightCNN-9 for CASIA NIR-VIS 2.0, Oulu-CASIA NIR-VIS and BUAA-VisNir, respectively. Furthermore, imposing the correlation alignment constraint in Eq. (\ref{eq_linear_projection}) can also boost the performance, which indicates that the assumptions of modeling the light spectrum variations via correlation alignment is reasonable and effective.

As shown in Table \ref{tab:ablation_study}, the improvements benefiting from three parts, including disentangled variational part, mean discrepancy part and correlation alignment constraint, verifies that our DVR method can significantly reduce the domain discrepancy and alleviate overfitting even if the number of training samples is not large enough.

\subsection{Comparisons}
The performance of the proposed DVR method based on both LightCNN-9 and LightCNN-29 is compared with some recent  state-of-the-art HFR methods in Table \ref{tab:HFR} on the three datasets.    The compared state-of-the-art HFR methods include both traditional handcrafted feature-based methods as well as deep learning-based methods.  In particular, the performance of handcrafted feature-based methods, such as KDSR~\cite{Huang2013RegularizedDS}, H2(LBP3)~\cite{Shao2017CrossModalityFL}, Gabor+RBM~\cite{Yi2014SharedRL}, Recon.+UDP~\cite{JuefeiXu2015NIRVISHF}, Gabor+JB~\cite{Chen2012BayesianFR} and Gabor+HJB~\cite{Shi2017CrossModalityFR}, as well as deep learning-based methods including IDNet~\cite{Reale2016SeeingTF}, HFR-CNN~\cite{Saxena2016HeterogeneousFR}, Hallucination~\cite{Lezama2017NotAO}, TRIVET~\cite{XXLiu:2016}, IDR~\cite{RHe:2017}, ADFL~\cite{Song2018AdversarialDH}, CDL~\cite{XWu:2018} and W-CNN~\cite{DBLP:journals/corr/abs-1708-02412} are compared in Table \ref{tab:HFR}.

For the most challenging CASIA NIR-VIS 2.0 database, it can be observed from the Table \ref{tab:HFR} that DVR performs better than the other compared methods. For fair comparisons, DVR on LightCNN-9 obtains \bm{$99.1\%$} on Rank-1 accuracy and \bm{$98.6\%$} on VR@FAR=0.1\%, which outperforms the other state-of-the-art methods on LightCNN-9, including TRIVET~\cite{XXLiu:2016}, IDR~\cite{RHe:2017}, ADFL~\cite{Song2018AdversarialDH}, CDL~\cite{XWu:2018} and W-CNN~\cite{DBLP:journals/corr/abs-1708-02412}. When the backbone is changed to LightCNN-29, DVR further gains \bm{$0.8\%$} on Rank-1 accuracy and \bm{$1.0\%$} on VR@FAR=0.1\%. The experimental results suggest that the domain discrepancy between NIR and VIS can be reduced by DVR.

For the Oulu-CASIA NIR-VIS and BUAA-VisNir databases, since the number of samples in the training set are not large enough, Table \ref{tab:HFR} demonstrates that benefiting from disentangled latent variables modeling, DVR outperforms previous state-of-the-art method such as W-CNN~\cite{DBLP:journals/corr/abs-1708-02412} by a large margin ($89.7\%$ vs $81.5\%$ on VR@FAR=1\% on Oulu-CASIA NIR-VIS as well as $97.0\%$ vs $96.0\%$ on VR@FAR=1\% on the BUAA-VisNir database). LightCNN-29, further improves the VR@FAR=0.1\% performance by \bm{$19.1\%$} and \bm{$4.1\%$} on the Oulu-CASIA NIR-VIS and BUAA-VisNir databases, respectively.

Fig. \ref{roc} shows the ROC curves corresponding to TRIVET~\cite{XXLiu:2016}, IDR~\cite{RHe:2017}, ADFL~\cite{Song2018AdversarialDH}, CDL~\cite{XWu:2018}, W-CNN~\cite{DBLP:journals/corr/abs-1708-02412}, DVR(LightCNN-9) and DVR(LightCNN-29). It can be observed that the ROC curves corresponding to the DVR method based on both LightCNN-9 and LightCNN-29 are significantly better than all the other methods. Again, this clearly shows the significance of the proposed framework for HFR.  When the False Positive Rate is larger than 0.01, the True Positive Rates of all the methods are close. When  the False Positive Rate tends to be small, there are large gaps between the curves of DVR and others.

\section{Conclusion}
A framework to disentangle the NIR and VIS heterogeneous face representations, called Disentangled Variational Representation (DVR), was proposed in this paper. It provides a novel way to disentangle the NIR and VIS representations with the identity information and their within-person variations. A variational lower bound is used to estimate the posterior and optimize the disentangled latent variable space. The minimization of the identity information for the same subject and the correlation alignment constraint on the modality variations further improve the representative ability of the disentangled latent variable. An alternative optimization is employed to provide mutual promotion for both disentangled variational representation and HFR network. In this way, we can easily generate NIR and VIS samples from the likelihood according to the disentangled representations, which can effectively alleviate overfitting for HFR on the limited number of training data. Experimental results demonstrate that the proposed DVR framework leads to excellent matching accuracy on three challenging HFR databases.  In addition, an ablation study is developed to demonstrate the improvements obtained by the different modules of the proposed framework.

\section{Acknowledgments}

This work is funded by National Natural Science Foundation of China (Grants No. 61622310) and Youth Innovation Promotion Association CAS (2015190).

{
\bibliographystyle{aaai}
\bibliography{egbib}

\begin{thebibliography}{}

\bibitem[\protect\citeauthoryear{Chen \bgroup et al\mbox.\egroup
  }{2009}]{JChen:2009}
Chen, J.; Yi, D.; Yang, J.; Zhao, G.; Li, S.~Z.; and Pietikainen, M.
\newblock 2009.
\newblock Learning mappings for face synthesis from near infrared to visual
  light images.
\newblock In {\em CVPR}.

\bibitem[\protect\citeauthoryear{Chen \bgroup et al\mbox.\egroup
  }{2012}]{Chen2012BayesianFR}
Chen, D.; Cao, X.; Wang, L.; Wen, F.; and Sun, J.
\newblock 2012.
\newblock Bayesian face revisited: A joint formulation.
\newblock In {\em ECCV}.

\bibitem[\protect\citeauthoryear{Desjardins, Courville, and
  Bengio}{2012}]{Desjardins2012DisentanglingFO}
Desjardins, G.; Courville, A.~C.; and Bengio, Y.
\newblock 2012.
\newblock Disentangling factors of variation via generative entangling.
\newblock {\em CoRR} abs/1210.5474.

\bibitem[\protect\citeauthoryear{Di, Zhang, and Patel}{2018}]{Patel_BTAS2018}
Di, X.; Zhang, H.; and Patel, V.~M.
\newblock 2018.
\newblock Polarimetric thermal to visible face verification via attribute
  preserved synthesis.
\newblock In {\em BTAS}.

\bibitem[\protect\citeauthoryear{Goodfellow \bgroup et al\mbox.\egroup
  }{2014}]{Goodfellow2014GenerativeAN}
Goodfellow, I.~J.; Pouget-Abadie, J.; Mirza, M.; Xu, B.; Warde-Farley, D.;
  Ozair, S.; Courville, A.~C.; and Bengio, Y.
\newblock 2014.
\newblock Generative adversarial networks.
\newblock In {\em NIPS}.

\bibitem[\protect\citeauthoryear{Goswami \bgroup et al\mbox.\egroup
  }{2011}]{Goswami2011EvaluationOF}
Goswami, D.; Chan, C.-H.; Windridge, D.; and Kittler, J.
\newblock 2011.
\newblock Evaluation of face recognition system in heterogeneous environments
  (visible vs nir).
\newblock {\em ICCV Workshops}.

\bibitem[\protect\citeauthoryear{Guo \bgroup et al\mbox.\egroup
  }{2016}]{DBLP:journals/corr/GuoZHHG16}
Guo, Y.; Zhang, L.; Hu, Y.; He, X.; and Gao, J.
\newblock 2016.
\newblock Ms-celeb-1m: {A} dataset and benchmark for large-scale face
  recognition.
\newblock In {\em ECCV}.

\bibitem[\protect\citeauthoryear{He \bgroup et al\mbox.\egroup
  }{2017}]{RHe:2017}
He, R.; Wu, X.; Sun, Z.; and Tan, T.
\newblock 2017.
\newblock Learning invariant deep representation for {NIR}-{VIS} face
  recognition.
\newblock In {\em AAAI}.

\bibitem[\protect\citeauthoryear{He \bgroup et al\mbox.\egroup
  }{2018}]{DBLP:journals/corr/abs-1708-02412}
He, R.; Wu, X.; Sun, Z.; and Tan, T.
\newblock 2018.
\newblock Wasserstein {CNN:} learning invariant features for {NIR}-{VIS} face
  recognition.
\newblock {\em TPAMI}.

\bibitem[\protect\citeauthoryear{Hu \bgroup et al\mbox.\egroup
  }{2018}]{hu2018pose}
Hu, Y.; Wu, X.; Yu, B.; He, R.; and Sun, Z.
\newblock 2018.
\newblock Pose-guided photorealistic face rotation.
\newblock In {\em CVPR}.

\bibitem[\protect\citeauthoryear{Huang \bgroup et al\mbox.\egroup
  }{2013}]{Huang2013RegularizedDS}
Huang, X.; Lei, Z.; Fan, M.; Wang, X.; and Li, S.~Z.
\newblock 2013.
\newblock Regularized discriminative spectral regression method for
  heterogeneous face matching.
\newblock {\em IEEE TIP}.

\bibitem[\protect\citeauthoryear{Huang \bgroup et al\mbox.\egroup
  }{2017}]{Huang2017BeyondFR}
Huang, R.; Zhang, S.; Li, T.; and He, R.
\newblock 2017.
\newblock Beyond face rotation: Global and local perception gan for
  photorealistic and identity preserving frontal view synthesis.
\newblock In {\em ICCV}.

\bibitem[\protect\citeauthoryear{Huang, Sun, and Wang}{2012}]{DHuang:2012}
Huang, D.; Sun, J.; and Wang, Y.
\newblock 2012.
\newblock The {BUAA}-{V}is{N}ir face database instructions.
\newblock Technical Report IRIP-TR-12-FR-001, Beihang University, Beijing,
  China.

\bibitem[\protect\citeauthoryear{Juefei-Xu, Pal, and
  Savvides}{2015}]{JuefeiXu2015NIRVISHF}
Juefei-Xu, F.; Pal, D.~K.; and Savvides, M.
\newblock 2015.
\newblock Nir-vis heterogeneous face recognition via cross-spectral joint
  dictionary learning and reconstruction.
\newblock {\em CVPR Workshops}.

\bibitem[\protect\citeauthoryear{Kan, Shan, and
  Chen}{2016}]{Kan2016MultiviewDN}
Kan, M.; Shan, S.; and Chen, X.
\newblock 2016.
\newblock Multi-view deep network for cross-view classification.
\newblock {\em CVPR}.

\bibitem[\protect\citeauthoryear{Kim and Mnih}{2018}]{Kim2018DisentanglingBF}
Kim, H., and Mnih, A.
\newblock 2018.
\newblock Disentangling by factorising.
\newblock In {\em ICML}.

\bibitem[\protect\citeauthoryear{Kingma and Ba}{2015}]{Kingma2014AdamAM}
Kingma, D.~P., and Ba, J.
\newblock 2015.
\newblock Adam: A method for stochastic optimization.
\newblock In {\em ICLR}.

\bibitem[\protect\citeauthoryear{Kingma and
  Welling}{2014}]{Kingma2013AutoEncodingVB}
Kingma, D.~P., and Welling, M.
\newblock 2014.
\newblock Auto-encoding variational bayes.
\newblock In {\em ICLR}.

\bibitem[\protect\citeauthoryear{Klare, Li, and
  Jain}{2011}]{Klare2011MatchingFS}
Klare, B.; Li, Z.; and Jain, A.~K.
\newblock 2011.
\newblock Matching forensic sketches to mug shot photos.
\newblock {\em IEEE TPAMI}.

\bibitem[\protect\citeauthoryear{Lei \bgroup et al\mbox.\egroup
  }{2007}]{ZLei:2007}
Lei, Z.; Chu, R.; He, R.; Liao, S.; and Li, S.~Z.
\newblock 2007.
\newblock Face recognition by discriminant analysis with gabor tensor
  representation.
\newblock In {\em ICB}.

\bibitem[\protect\citeauthoryear{Lezama, Qiu, and
  Sapiro}{2017}]{Lezama2017NotAO}
Lezama, J.; Qiu, Q.; and Sapiro, G.
\newblock 2017.
\newblock Not afraid of the dark: Nir-vis face recognition via cross-spectral
  hallucination and low-rank embedding.
\newblock {\em CVPR}.

\bibitem[\protect\citeauthoryear{Li \bgroup et al\mbox.\egroup
  }{2013}]{SLi:2013}
Li, S.~Z.; Yi, D.; Lei, Z.; and Liao, S.
\newblock 2013.
\newblock The casia {nir}-{vis} 2.0 face database.
\newblock In {\em CVPR Workshops}.

\bibitem[\protect\citeauthoryear{Liao \bgroup et al\mbox.\egroup
  }{2009}]{Liao2009HeterogeneousFR}
Liao, S.; Yi, D.; Lei, Z.; Qin, R.; and Li, S.~Z.
\newblock 2009.
\newblock Heterogeneous face recognition from local structures of normalized
  appearance.
\newblock In {\em ICB}.

\bibitem[\protect\citeauthoryear{Lin and Tang}{2006}]{Lin2006IntermodalityFR}
Lin, D., and Tang, X.
\newblock 2006.
\newblock Inter-modality face recognition.
\newblock In {\em ECCV}.

\bibitem[\protect\citeauthoryear{Liu \bgroup et al\mbox.\egroup
  }{2016}]{XXLiu:2016}
Liu, X.; Song, L.; Wu, X.; and Tan, T.
\newblock 2016.
\newblock Transferring deep representation for nir-vis heterogeneous face
  recognition.
\newblock In {\em ICB}.

\bibitem[\protect\citeauthoryear{Matthey \bgroup et al\mbox.\egroup
  }{2017}]{Matthey2016vaeLB}
Matthey, L.; Pal, A.; Burgess, C.; Glorot, X.; Botvinick, M.; Mohamed, S.; and
  Lerchner, A.
\newblock 2017.
\newblock $\beta$-{VAE}: Learning basic visual concepts with a constrained
  variational framework.
\newblock In {\em ICLR}.

\bibitem[\protect\citeauthoryear{Ouyang \bgroup et al\mbox.\egroup
  }{2016}]{Ouyang2016ASO}
Ouyang, S.; Hospedales, T.~M.; Song, Y.-Z.; and Li, X.
\newblock 2016.
\newblock A survey on heterogeneous face recognition: Sketch, infra-red, 3d and
  low-resolution.
\newblock {\em IVC}.

\bibitem[\protect\citeauthoryear{Parkhi, Vedaldi, and
  Zisserman}{2015}]{Parkhi:2015}
Parkhi, O.~M.; Vedaldi, A.; and Zisserman, A.
\newblock 2015.
\newblock Deep face recognition.
\newblock In {\em BMVC}.

\bibitem[\protect\citeauthoryear{Reale \bgroup et al\mbox.\egroup
  }{2016}]{Reale2016SeeingTF}
Reale, C.; Nasrabadi, N.~M.; Kwon, H.; and Chellappa, R.
\newblock 2016.
\newblock Seeing the forest from the trees: A holistic approach to
  near-infrared heterogeneous face recognition.
\newblock {\em CVPR Workshop}.

\bibitem[\protect\citeauthoryear{Saxena and
  Verbeek}{2016}]{Saxena2016HeterogeneousFR}
Saxena, S., and Verbeek, J.
\newblock 2016.
\newblock Heterogeneous face recognition with cnns.
\newblock In {\em ECCV Workshop}.

\bibitem[\protect\citeauthoryear{Schmidhuber}{1992}]{Schmidhuber1992LearningFC}
Schmidhuber, J.
\newblock 1992.
\newblock Learning factorial codes by predictability minimization.
\newblock {\em Neural Computation}.

\bibitem[\protect\citeauthoryear{Shao and Fu}{2017}]{Shao2017CrossModalityFL}
Shao, M., and Fu, Y.
\newblock 2017.
\newblock Cross-modality feature learning through generic hierarchical
  hyperlingual-words.
\newblock {\em IEEE TNNLS}.

\bibitem[\protect\citeauthoryear{Shao, Kit, and
  Fu}{2014}]{Shao2014GeneralizedTS}
Shao, M.; Kit, D.; and Fu, Y.
\newblock 2014.
\newblock Generalized transfer subspace learning through low-rank constraint.
\newblock {\em IJCV}.

\bibitem[\protect\citeauthoryear{Shi \bgroup et al\mbox.\egroup
  }{2017}]{Shi2017CrossModalityFR}
Shi, H.; Wang, X.; Yi, D.; Lei, Z.; Zhu, X.; and Li, S.~Z.
\newblock 2017.
\newblock Cross-modality face recognition via heterogeneous joint bayesian.
\newblock {\em IEEE SPL}.

\bibitem[\protect\citeauthoryear{Song \bgroup et al\mbox.\egroup
  }{2018}]{Song2018AdversarialDH}
Song, L.; Zhang, M.; Wu, X.; and He, R.
\newblock 2018.
\newblock Adversarial discriminative heterogeneous face recognition.
\newblock In {\em AAAI}.

\bibitem[\protect\citeauthoryear{Sun and Saenko}{2016}]{BSun:2016}
Sun, B., and Saenko, K.
\newblock 2016.
\newblock Deep coral: Correlation alignment for deep domain adaptation.
\newblock In {\em ECCV Workshops}.

\bibitem[\protect\citeauthoryear{Tang and Wang}{2003}]{Tang2003FaceSS}
Tang, X., and Wang, X.
\newblock 2003.
\newblock Face sketch synthesis and recognition.
\newblock In {\em ICCV}.

\bibitem[\protect\citeauthoryear{Wang and Tang}{2009}]{Wang2009FacePS}
Wang, X., and Tang, X.
\newblock 2009.
\newblock Face photo-sketch synthesis and recognition.
\newblock {\em IEEE TPAMI}.

\bibitem[\protect\citeauthoryear{Wang \bgroup et al\mbox.\egroup
  }{2015}]{Wang2015OnDM}
Wang, W.; Arora, R.; Livescu, K.; and Bilmes, J.~A.
\newblock 2015.
\newblock On deep multi-view representation learning.
\newblock In {\em ICML}.

\bibitem[\protect\citeauthoryear{Wang \bgroup et al\mbox.\egroup
  }{2016}]{Wang2016JointFS}
Wang, K.; He, R.; Wang, L.; Wang, W.; and Tan, T.
\newblock 2016.
\newblock Joint feature selection and subspace learning for cross-modal
  retrieval.
\newblock {\em IEEE TPAMI}.

\bibitem[\protect\citeauthoryear{Wu \bgroup et al\mbox.\egroup
  }{2018a}]{Wu2018ALC}
Wu, X.; He, R.; Sun, Z.; and Tan, T.
\newblock 2018a.
\newblock A light cnn for deep face representation with noisy labels.
\newblock {\em IEEE TIFS}.

\bibitem[\protect\citeauthoryear{Wu \bgroup et al\mbox.\egroup
  }{2018b}]{XWu:2018}
Wu, X.; Song, L.; He, R.; and Tan, T.
\newblock 2018b.
\newblock Coupled deep learning for heterogeneous face recognition.
\newblock In {\em AAAI}.

\bibitem[\protect\citeauthoryear{Xiao \bgroup et al\mbox.\egroup
  }{2013}]{Xiao2013CoupledFS}
Xiao, L.; Sun, Z.; He, R.; and Tan, T.
\newblock 2013.
\newblock Coupled feature selection for cross-sensor iris recognition.
\newblock In {\em BTAS}.

\bibitem[\protect\citeauthoryear{Yi \bgroup et al\mbox.\egroup
  }{2015}]{Yi2014SharedRL}
Yi, D.; Lei, Z.; Liao, S.; and Li, S.~Z.
\newblock 2015.
\newblock Shared representation learning for heterogeneous face recognition.
\newblock In {\em FG Workshops}.

\bibitem[\protect\citeauthoryear{Zhang \bgroup et al\mbox.\egroup
  }{2017}]{Patel_IJCB2017}
Zhang, H.; Patel, V.~M.; Riggan, B.~S.; and Hu, S.
\newblock 2017.
\newblock Generative adversarial network-based synthesis of visible faces from
  polarimetrie thermal faces.
\newblock In {\em IJCB}.

\bibitem[\protect\citeauthoryear{Zhao \bgroup et al\mbox.\egroup
  }{2017}]{Zhao2017DualAgentGF}
Zhao, J.; Xiong, L.; Karlekar, J.; Li, J.; Zhao, F.; Wang, Z.; Pranata, S.;
  Shen, S.; Yan, S.; and Feng, J.
\newblock 2017.
\newblock Dual-agent gans for photorealistic and identity preserving profile
  face synthesis.
\newblock In {\em NIPS}.

\bibitem[\protect\citeauthoryear{Zhu \bgroup et al\mbox.\egroup
  }{2014}]{Zhu2014MatchingNF}
Zhu, J.-Y.; Zheng, W.-S.; Lai, J.; and Li, S.~Z.
\newblock 2014.
\newblock Matching nir face to vis face using transduction.
\newblock {\em IEEE TIFS}.

\bibitem[\protect\citeauthoryear{Zhu \bgroup et al\mbox.\egroup
  }{2017}]{Zhu2017UnpairedIT}
Zhu, J.-Y.; Park, T.; Isola, P.; and Efros, A.~A.
\newblock 2017.
\newblock Unpaired image-to-image translation using cycle-consistent
  adversarial networks.
\newblock {\em ICCV}.

\end{thebibliography}
}

.

\end{document}